\documentclass[a4paper,12pt,onecolumn]{article}

\usepackage{FAS}
\usepackage[utf8]{inputenc}
\usepackage{graphicx}
\usepackage{booktabs}
\usepackage{multirow}
\usepackage[table]{xcolor}
\usepackage{tabularx}
\usepackage{amsmath}
\usepackage{amssymb}
\usepackage[hidelinks]{hyperref}
\usepackage{cleveref}
\usepackage{enumitem}
\usepackage{array}

\usepackage{fontawesome5}
  \usepackage{tikz}
  \usepackage{pgfplots}
\usepackage{colortbl}
\usepackage{makecell}
\usepackage{subcaption}
\usepackage{siunitx}
\usepackage[most]{tcolorbox}

\definecolor{guidecolor}{RGB}{0, 100, 180}
\definecolor{todocolor}{RGB}{200, 80, 0}
\definecolor{examplecolor}{RGB}{100, 150, 50}
\definecolor{KIT-green}{RGB}{0,150,130}

\newtcolorbox{kitscenes}[1][KITScenes Case Study]{
  enhanced,
  colback=KIT-green!4!white,
  colframe=KIT-green,
  coltitle=white,
  fonttitle=\bfseries\small,
  title={\faMapMarker*~#1},
  boxrule=0.8pt, arc=1.5mm,
  left=2mm, right=2mm, top=1.5mm, bottom=1.5mm,
  before skip=6pt, after skip=6pt,
}

\begin{document}

\title{Creating Impactful Autonomous Driving Datasets:\\
 A Strategic Guide from Research Gap to Benchmark}

\author{\small%
  \parbox{0.96\textwidth}{\centering
Richard Schwarzkopf$^\dagger$\thanks{all authors are associated with Karlsruhe Institute of Technology (KIT), all dataset authors are ordered alphabetically, $^\dagger$also associated with the FZI Research Center for Information Technology, Karlsruhe. E-Mail: (firstname.lastname@kit.edu).},
Jonas Merkert,
Frank Bieder$^\dagger$,\\
Annika Bätz,
Alexander Blumberg,
Carlos Fernandez,
Felix Hauser$^\dagger$,
Fabian Immel$^\dagger$,
Christian Kinzig,
Hendrik Königshof$^\dagger$,
Fabian Konstantinidis,
Martin Lauer,
Willi Poh,
Nils Rack,
Kevin Rösch$^\dagger$,
Yinzhe Shen,
Marlon Steiner,
Gleb Stepanov,
Dominik Strutz,
Ömer Şahin Taş,
Julian Truetsch$^\dagger$,
Kaiwen Wang,
Royden Wagner,\\
Jan-Hendrik Pauls,
Christoph Stiller$^\dagger$}
}
\date{}
\maketitle \thispagestyle{empty}

\begin{abstract}

Well-designed autonomous driving datasets have fundamentally shaped research progress, yet existing literature primarily describes \textit{what} datasets contain rather than \textit{how} to strategically design impactful ones. This is especially limiting for small and medium-sized labs and startups that cannot afford to misallocate scarce resources. We argue that impactful dataset creation begins with a diagnosis: whether a research question is blocked by a \emph{data problem} or an \emph{evaluation problem}, and proceeds by selecting the \emph{minimal data operator(s)} that closes the resulting gap, recording new data only when no cheaper operator(s) suffices. We analyze the evolution of major autonomous driving (AD) datasets through this lens and distill a strategic framework spanning gap identification, operator choice, sensor suite design, and annotation strategy. We ground the framework in a running case study of our KITScenes dataset family.
The datasets are available at: \href{https://kitscenes.com/}{https://kitscenes.com/}.
\end{abstract}


\section{Introduction}
\label{sec:introduction}

In autonomous driving (AD) research, public datasets play a crucial role for developing and evaluating new methods, enabling fair comparison of methods. However, existing literature often only describes \emph{what} a dataset contains, providing an ex-post-facto account of the sensor modalities, the geographic diversity, or the provided annotations. Far fewer works address \emph{how} to strategically design an impactful dataset or \emph{why} certain decisions were made, offering little guidance on systematically identifying high-impact research gaps, navigating cost trade-offs, and determining the best approach to advance the field in both the short and long term.

We argue that impactful dataset creation should be driven by three questions asked \emph{before} any data is collected. First, a diagnosis: What is the main research gap the team is most interested in, and where are they confident they can close this gap significantly?

Second, is the targeted research question blocked because a method \emph{cannot be developed} for lack of training data (a \textbf{data problem}), or because its performance \emph{cannot be conclusively measured and compared} (an \textbf{evaluation problem})?
Third, a choice of means: among the available \emph{data operators}, from re-labeling and combining existing datasets, through synthesis (simulation or generation), to recording new real-world data, which is the \emph{minimal} one that closes the diagnosed gap?

\medskip
To make this concrete, we present a running case study of our \textbf{KITScenes} dataset family, which deliberately spans the two extremes of the cost spectrum: \emph{KITScenes Multimodal}~\cite{kitscenes_mm}, a high-fidelity recorded dataset with production-grade HD maps, where recording was the justified operator; and \emph{KITScenes LongTail}~\cite{wagner2026longtail}, a lean long-tail evaluation benchmark that pairs low-cost recording with high-value human reasoning traces. Notably, KITScenes Multimodal had to defend itself against the option to \emph{extend} Argoverse~2~\cite{Argoverse2} with new map labels, a far cheaper operator, and we use that abandoned decision to illustrate when recording does, and does not, pay off. Our contributions are:
\begin{itemize}
    \item An analysis of major AD datasets through the lens of the research gap each addressed and the \emph{data operator} used to create it (\Cref{sec:evolution}).
    \item A strategic framework that guides a researcher from diagnosing a data vs.\ evaluation problem, through choosing the minimal data operator, to sensor, annotation, and scaling decisions (\Cref{sec:framework}).
    \item A running case study of the KITScenes family that grounds each step in concrete decisions, including the record-vs-reuse trade-off, and distills the lessons we learned.
\end{itemize}

\section{The Evolution of AD Datasets}
\label{sec:evolution}

The role of datasets in robotics and autonomous driving has shifted fundamentally over the past decade. Early benchmarks were conceived primarily as instruments to rigorously and reproducibly \emph{evaluate} systems and methods. With the rise of powerful machine learning, the emphasis drifted from data as an evaluation basis toward data as the engine of data-driven development. This shift kept growing in scale and application diversity, and has recently produced a divergence between the data used for \emph{training} and the data used for \emph{evaluation}: internet-scale pretrained models are adapted via in-context learning or fine-tuned on large-scale data such as the NVIDIA Physical AI dataset \cite{NvidiaAD} or nuPlan \cite{nuPlan}, while being evaluated on curated, long-tail benchmarks such as the Waymo Open Dataset for End-to-End Driving \cite{WaymoE2E}. This split mirrors our central distinction between the \emph{data problem} (training) and the \emph{evaluation problem} (benchmarking), developed in \Cref{sec:framework}. This section surveys how AD datasets evolved along this trajectory: \Cref{tab:evolution} summarizes the key datasets and, crucially, the \emph{data operator} each used, while \Cref{fig:av_datasets} illustrates the resulting separation into large-scale training data and high-fidelity evaluation data.
We organize this evolution into clusters by era, data operator (the way the data was produced, defined in \Cref{subsec:operators}), and annotation regime (\Cref{tab:evolution}), describing each by the research it enabled rather than cataloguing datasets.
The \textbf{foundation} cluster (2012-2017) established that progress needs standardized, reproducible evaluation: KITTI \cite{Kitti} introduced recorded, human-verified multi-sensor data with held-out splits, enabling core detection, tracking, flow, depth, and odometry research.
In parallel, a \textbf{synthetic} cluster used simulation for cheap pixel-perfect labels and controllable domains, from photo-realistic clones \cite{virtualkitti} to engine-based closed-loop testing.
The \textbf{multimodal robotaxi} cluster (2018-2022) then prioritized scale and sensor fusion: nuScenes \cite{nuscenes} first delivered a full 360\textdegree{} suite with HD maps, and Argoverse~2 \cite{Argoverse2} and the Waymo Open Dataset \cite{WaymoOpenPerception, WaymoOpenMotion} scaled multimodal detection, tracking, forecasting, and online/changed-map research.
A distinct \textbf{crowdsourced collection} cluster instead mined commodity-device imagery (dashcams, phones) \cite{bdd100k, neuhold2017mapillary}, trading fidelity for geographic and appearance diversity to drive robustness and domain generalization.
More recently, \textbf{derived and unified} benchmarks \cite{openlaneV2, wild2025argotweak, Dauner2026123D} reused existing data to define new tasks and reduce fragmentation without new recording.
Finally, the field split into two regimes (\Cref{fig:av_datasets}): massive auto-labeled and generative data for planning and world models \cite{nuPlan, NvidiaAD, nvidia2025cosmosdrivedreams}, versus small, high-fidelity sets for conclusive long-tail evaluation \cite{WaymoE2E, wagner2026longtail} critical for Level~4 autonomy.

\pgfdeclareplotmark{pizza}{%
  \pgfpathmoveto{\pgfpointorigin}%
  \pgfpathlineto{\pgfpointpolar{180}{\pgfplotmarksize}}%
  \pgfpatharc{180}{90}{\pgfplotmarksize}%
  \pgfpathclose%
  \pgfusepath{fill,stroke}%
}

\begin{figure}[htbp]
  \centering
  \begin{tikzpicture}
    \begin{semilogyaxis}[
      width  = 0.92\linewidth,
      height = 7.9cm,
      xlabel = {Year},
      ylabel = {Driving Hours},
      xmin   = 2011,
      xmax   = 2027,
      ymin   = 0.4,
      ymax   = 5000,
      xtick  = {2012, 2014, 2016, 2018, 2020, 2022, 2024, 2026},
      xticklabels = {2012, 2014, 2016, 2018, 2020, 2022, 2024, 2026},
      x tick label style = {rotate=45, anchor=east, font=\small},
      ytick  = {1, 10, 100, 1000},
      yticklabels = {1, 10, 100, 1000},
      ymajorgrids      = true,
      yminorgrids      = false,
      major grid style = {line width=0.25pt, draw=gray!20},
      minor tick num   = 0,
      legend pos        = north west,
      legend style      = {font=\tiny, draw=gray!50, fill=white,
                           fill opacity=0.9, text opacity=1,
                           inner sep=2pt, row sep=0pt,
                           legend image post style={scale=0.7}},
      legend cell align = left,
      clip = false,
    ]

    \addplot[
      only marks,
      mark        = *,
      mark size   = 4pt,
      color       = blue!70!black,
    ] coordinates { (2012, 1.5) };
    \addlegendentry{KITTI}

    \addplot[
      only marks,
      mark        = square*,
      mark size   = 3.5pt,
      color       = orange!80!black,
    ] coordinates { (2019, 5.5) };
    \addlegendentry{nuScenes}

    \addplot[
      only marks,
      mark        = triangle*,
      mark size   = 4pt,
      color       = green!60!black,
    ] coordinates { (2019, 0.8) };
    \addlegendentry{Argoverse\,1}

    \addplot[
      only marks,
      mark        = star,
      mark size   = 4pt,
      color       = magenta!70!black,
    ] coordinates { (2019, 2.5) };
    \addlegendentry{Lyft Level\,5}

    \addplot[
      only marks,
      mark        = diamond*,
      mark size   = 4pt,
      color       = red!75!black,
    ] coordinates { (2020, 11) };
    \addlegendentry{Waymo Perception}

    \addplot[
      only marks,
      mark        = pentagon*,
      mark size   = 4pt,
      color       = violet!80!black,
    ] coordinates { (2021, 144) };
    \addlegendentry{ONCE}

    \addplot[
      only marks,
      mark        = *,
      mark size   = 4pt,
      color       = cyan!60!black,
    ] coordinates { (2022, 1.5) };
    \addlegendentry{KITTI-360}

    \addplot[
      only marks,
      mark        = square*,
      mark size   = 3.5pt,
      color       = brown!70!black,
    ] coordinates { (2022, 4.2) };
    \addlegendentry{Argoverse\,2 Sensor}

    \addplot[
      only marks,
      mark        = triangle*,
      mark size   = 4pt,
      color       = teal!70!black,
    ] coordinates { (2022, 120) };
    \addlegendentry{nuPlan Sensor}

    \addplot[
      only marks,
      mark        = triangle*,
      mark size   = 4pt,
      color       = teal,
    ] coordinates { (2022, 1200) };
    \addlegendentry{nuPlan Full}
    
    \addplot[
    only marks,
    mark        = diamond*,
    mark size   = 4pt,
    color       = red!50!black,
    ] coordinates { (2025, 12) };
    \addlegendentry{WOD-E2E}
    
    \addplot[
      only marks,
      mark        = diamond*,
      mark size   = 4.5pt,
      color       = black,
    ] coordinates { (2026, 1700) };
    \addlegendentry{NVIDIA PhysAI AV}

    \addplot[
      only marks,
      mark        = pizza,
      mark size   = 6pt,
      color       = KIT-green!60,
    ] coordinates { (2026, 2.5) };
    \addlegendentry{KITScenes LongTail}
    
    \addplot[
      only marks,
      mark        = pizza,
      mark size   = 6pt,
      color       = KIT-green,
    ] coordinates { (2026, 6) };
    \addlegendentry{KITScenes Multimodal}

    \node[font=\tiny, gray, rotate=0, anchor=south]
    at (axis cs:2022.9, 500) {large-scale training data};
    
    \node[font=\tiny, gray, rotate=0, anchor=east]
    at (axis cs:2025.6, 2.5) {specialized eval~\&~FT~\&~LT};

    \node[font=\scriptsize, anchor=west, color=blue!70!black]
      at (axis cs:2012.1, 1.5)       {KITTI};
    \node[font=\scriptsize, anchor=east, color=orange!80!black]
      at (axis cs:2018.9, 5.5)     {nuScenes};
    \node[font=\scriptsize, anchor=north east, color=green!60!black]
      at (axis cs:2018.9, 1.4)     {AV1};
    \node[font=\scriptsize, anchor= east, color=magenta!70!black]
      at (axis cs:2019, 2.1)       {Lyft};
    \node[font=\scriptsize, anchor=south west, color=red!75!black]
      at (axis cs:2020.1, 6.4)     {WOD};
    \node[font=\scriptsize, anchor=west, color=violet!80!black]
      at (axis cs:2019.4, 144)     {ONCE};
    \node[font=\scriptsize, anchor=north, color=cyan!60!black]
      at (axis cs:2022, 1.3)       {K-360};
    \node[font=\scriptsize, anchor=south west, color=brown!70!black]
      at (axis cs:2022.1, 4.2)     {AV2};
    \node[font=\scriptsize, anchor=south west, color=teal!70!black]
      at (axis cs:2022.1, 120)     {nuPlan Sensor};
    \node[font=\scriptsize, anchor=south east, color=teal]
      at (axis cs:2021.9, 1200)     {nuPlan Full};
    \node[font=\scriptsize, anchor=west, color=black]
      at (axis cs:2024.1, 1700)    {NVIDIA};
    \node[font=\scriptsize, anchor=south east, color=red!50!black]
      at (axis cs:2025.1, 12) {WOD-E2E};
    \node[font=\scriptsize, anchor=south east, color=KIT-green]
      at (axis cs:2025.8, 5)    {KS-MM};
    \node[font=\scriptsize, anchor=north, color=KIT-green!60]
      at (axis cs:2026, 2.1) {KS-LT};
    \end{semilogyaxis}
  \end{tikzpicture}
  \caption{Publicly available autonomous driving dataset with sensor data and labels (no KITTI raw data recording, NVIDIA labels not public yet but promised, WOD with current size). A scale separation trend can be observed for datasets designed for scale with autolabeling and datasets designed for high fidelity and specialized evaluation, fine-tuning, or long-tail distribution sampling. 
           }
  \label{fig:av_datasets}
\end{figure}
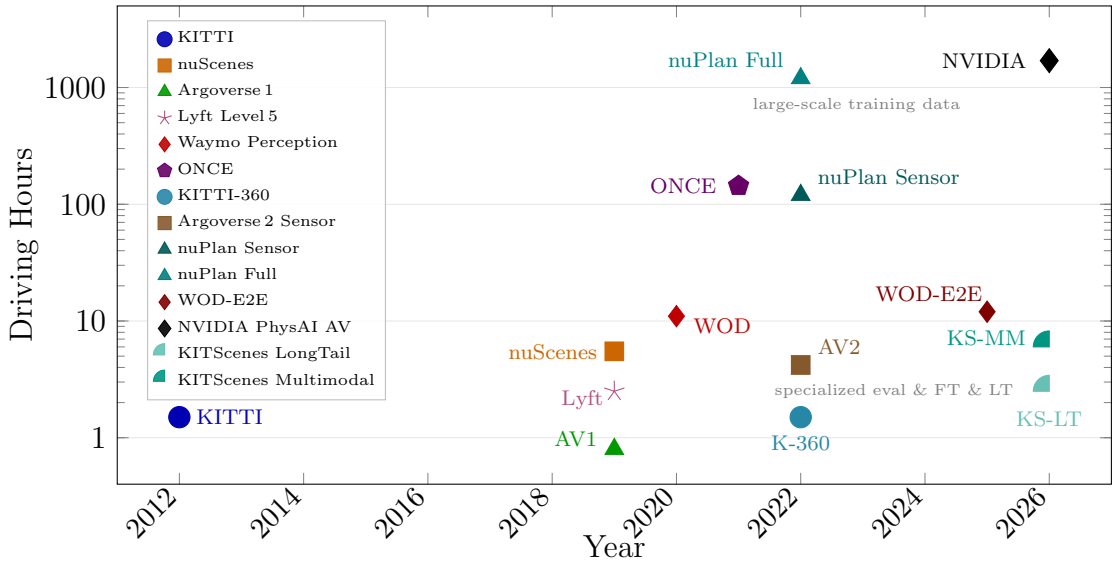

\noindent
\textbf{Emerging needs:}
Despite the wealth of available data, this evolution has revealed requirements that current benchmarks only partially address. First, the training/evaluation divergence raises the bar for curated evaluation sets: as models are increasingly trained on large-scale auto-labeled data, conclusive benchmarking demands the highest sensor, calibration, synchronization, and annotation fidelity and data sampled deliberately for long-tail coverage. Second, a persistent gap exists between research data and the real-world software stacks in which AV methods are ultimately deployed, both in data formats and annotation richness. A concrete instance is the lack of HD maps in standardized, open-source formats such as Lanelet2 \cite{poggenhans2018lanelet2}, with deployment-ready completeness including explicit logical associations between 3D traffic lights and the lanes they govern, which are essential for planning and standard in production maps, yet largely absent from public research datasets.
These emerging needs are the design target of our case study (\Cref{sec:framework}). The KITScenes family is built to close them on underrepresented European roads, and we use it throughout the remainder of this paper to ground the strategic framework.

\begin{table}[!h]
\def\arraystretch{1.2}
\centering
\caption{Evolution of autonomous driving datasets, grouped into \emph{clusters} by era, the \emph{data operator(s)} used and their dominant \emph{annotation regime}. Each row names only the most seminal dataset of the cluster; further examples are cited compactly. Operators (\textbf{R}, \textbf{C}, \textbf{U}, \textbf{E}, \textbf{S}) are defined in \Cref{tab:dataset_operators} (\Cref{subsec:operators}). Annotation: \textbf{H}\,=\,human-in-the-loop, \textbf{A}\,=\,auto-labeling.}
\label{tab:evolution}
\small
\resizebox{\linewidth}{!}{%
\begin{tabular}{>{\arraybackslash}p{2.6cm} c c c >{\arraybackslash}p{2.4cm} >{\arraybackslash}p{6.2cm} }
\toprule
\textbf{Cluster} & \textbf{Era} & \textbf{Op.} & \textbf{Ann.} & \textbf{Seminal (+ refs)} & \textbf{Gap $\rightarrow$ Enabled} \\
\midrule
Foundation perception benchmarks & 2012--17 & \textbf{R} & \textbf{H} & KITTI \cite{Kitti,Cordts2016Cityscapes,Liao2022PAMI,behley2019iccv,huang2018apolloscape} & First standardized real-world benchmarks $\rightarrow$ detection, tracking, flow, depth, odometry, segmentation. \\
Synthetic \& simulation & 2016-- & \textbf{S} & \textbf{A} & Virtual KITTI \cite{virtualkitti,dosovitskiy2017carla,jia2024bench2drive} & Cheap pixel-perfect labels $\rightarrow$ controllable domains, closed-loop training/eval. \\
Multimodal robotaxi fleets & 2018--22 & \textbf{R} & \textbf{H} & nuScenes \cite{nuscenes,WaymoOpenPerception,WaymoOpenMotion,Argoverse2,Argoverse,houston2021lyft,mao2021once,alibeigi2023zenseact,pandaset2021,geyer2020a2d2} & 360\textdegree{} suites + HD maps at scale $\rightarrow$ 3D detection/tracking, forecasting, online/changed maps. \\
Crowdsourced collection & 2017-- & \textbf{C} & \textbf{H+A} & BDD100K \cite{bdd100k,neuhold2017mapillary,schafer2018comma2k19} & Commodity-device diversity at low cost $\rightarrow$ robustness, domain generalization. \\
Derived \& unified benchmarks & 2023-- & \textbf{U/E} & \textbf{H} & OpenLane-V2 \cite{openlaneV2,wild2025argotweak,Dauner2026123D} & New tasks/formats from existing data $\rightarrow$ topology, map verification, cross-dataset training. \\
Large-scale planning \& E2E / generative & 2024-- & \textbf{R/S} & \textbf{A} & nuPlan \cite{nuPlan,NvidiaAD,nvidia2025cosmosdrivedreams} & Massive auto-labeled \& generative data $\rightarrow$ learning-based planning, world models. \\
Long-tail \& closed-loop evaluation & 2024-- & \textbf{R/E/S} & \textbf{H} & Waymo E2E \cite{WaymoE2E,Dauner2024NEURIPS,ljungbergh2024neuroncap} & Conclusive eval lagged training scale $\rightarrow$ long-tail, closed-loop E2E evaluation. \\
\midrule
\textbf{KITScenes Multimodal} & 2026 & \textbf{R} & \textbf{H} & KITScenes MM & High-fidelity eval $\rightarrow$ stack-ready Lanelet2 HD maps with traffic-light-to-lane links. \\
\textbf{KITScenes LongTail} & 2026 & \textbf{R} & \textbf{H} & KITScenes LT & Long-tail 360\textdegree{} eval $\rightarrow$ E2E logs with reasoning traces, ranking score. \\
\bottomrule
\end{tabular}
}
\end{table}

\section{Strategic Framework for Dataset and Evaluation Design}
\label{sec:framework}

We now turn the observations of \Cref{sec:evolution} into an actionable process. The framework is a sequence of decisions, each of which narrows the design space and constrains the next: (i)~\emph{diagnose} whether the research question is blocked by a data or an evaluation problem; (ii)~\emph{identify} the precise gap to close; (iii)~\emph{choose the minimal data operator} that closes it; and only then (iv)~commit to the concrete sensor, annotation, and scaling decisions that the chosen operator entails. The guiding principle throughout is impact-per-effort: each step is framed to spend scarce resources where they create the most leverage.

We illustrate every step with the running KITScenes case study, set apart in the green callout boxes so that the general guidance remains distinguishable from examples that are specific to our dataset family.

\subsection{Diagnose: Data Problem vs.\ Evaluation Problem}
\label{subsec:diagnose}

First, a researcher should be precise about \emph{why} progress is currently blocked. We find it clarifying to separate two failure modes. A \textbf{data problem} exists when a capable method \emph{cannot be developed or trained} because the data does not exist at sufficient scale or diversity; the bottleneck is on the supply side. An \textbf{evaluation problem} exists when methods can be built, but their performance \emph{cannot be conclusively measured, compared, or trusted}, e.g.\ because no benchmark captures the relevant conditions, the metrics are uninformative, or the test data is too easy, too small, or contaminated by leakage. The two are not mutually exclusive, but naming the dominant one is decisive, because it routes every subsequent choice: a data problem typically calls for scale or diversity and is often best served by cheaper operators (\Cref{subsec:operators}), whereas an evaluation problem calls for fidelity, careful curation, and metric design, and may justify the most expensive operator and annotation strategy.

This distinction also explains the field-wide divergence noted in \Cref{sec:evolution}: large auto-labeled corpora answer the data problem of training, while smaller, high-fidelity, long-tail sets answer the evaluation problem of trustworthy benchmarking. A common and costly mistake is to attack an evaluation problem with a scale-oriented data effort, or vice versa.

\begin{kitscenes}
KITScenes datasets target both \emph{data} (data from the EU with deployment-grade HD maps) and an \emph{evaluation} problem. Modern perception and end-to-end models can already be \emph{trained} on abundant data, yet they cannot be \emph{conclusively evaluated} for application-relevant capabilities: long-range perception, complete and stack-ready HD maps, and competent behavior in rare events. KITScenes Multimodal addresses the fidelity side (the highest sensor and map fidelity for conclusive benchmarking on irregular road layouts in dense European urban environments), while KITScenes LongTail addresses the coverage side (deliberate long-tail sampling).
\end{kitscenes}

\subsection{Identifying the Research Gap and Target Benchmarks}
\label{subsec:gaps}

High-impact datasets of the past identified their gap through close study of state-of-the-art methods and their lack of rigorous, large-scale evaluation.

\begin{figure}[tb]
    \centering
    \includegraphics[width=0.65\linewidth]{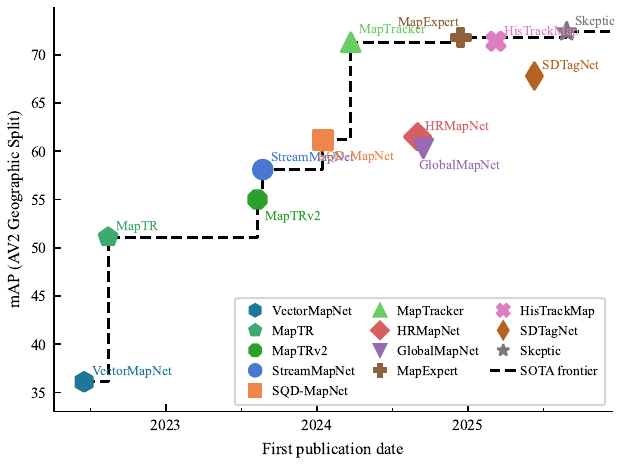}
    \caption{Historical SOTA progression of representative online HD map construction methods~\cite{maptr,maptrv2,yuan2024streammapnet,wang2024stream_sqd_mapnet,chen2024maptracker,zhang2024enhancing_HR_mapnet,shi2024globalmapnet,zhang2025mapexpert,yang2025histrackmap,immel2026sdtagnet,erdougan2025mapping_skeptic} on AV2~\cite{Argoverse2}. Progress saturates after MapTracker~\cite{chen2024maptracker}, signalling a gap of evaluation conclusiveness rather than of incremental accuracy. Figure referenced by~\cite{kitscenes_mm}.}
    \label{fig:history_online_hd_map_construction}
\end{figure}

We advocate approaching a dataset project with a specific target benchmark and task in mind from the outset, since this choice dictates the required sensors, annotations, and metrics. A candidate gap can be located along several complementary dimensions: the \emph{saturation} of an existing task, the definition of an \emph{entirely new task}, and the \emph{application alignment} of an established task.
We diagnosed in KITScenes Multimodal~\cite{kitscenes_mm} a concrete gap in online HD map construction. On existing benchmarks the task has visibly saturated as depicted in \Cref{fig:history_online_hd_map_construction}, while model outputs remain far from complete, missing most of the relational 3D map elements that deployment actually requires. This is a gap of conclusiveness and application alignment rather than of further incremental accuracy.

\begin{kitscenes}
KITScenes addresses two initial research gaps, one per dataset. \emph{KITScenes Multimodal} targets: no curated, high-fidelity dataset that pairs surround-view, long-range sensing with open-format, stack-ready HD maps that carry the logical associations (e.g.\ traffic-light-to-lane) required for planning. Closing it enables a more complete online HD map construction that recovers map elements \emph{and} their relations, and its conclusiveness rests on the highest sensor, calibration, and annotation fidelity together with relational map-construction metrics that keep detection and topology separable, unlike the aggregated OpenLane-V2 score~\cite{openlaneV2}. While this research gap is the initially targeted one, more research is enabled by its high-fidelity data and labels. \emph{KITScenes LongTail} targets: decision-making in long-tail scenarios cannot be conclusively measured, because matching a single expert trajectory by displacement error ignores that multiple maneuvers are valid. It frames reasoned, instruction-grounded long-tail driving as a new task and scores it with the Multi-Maneuver Score~\cite{wagner2026longtail} and human reasoning traces rather than displacement error alone. Together the two datasets span three design dimensions at once: \emph{new tasks}, \emph{application alignment}, and \emph{evaluation conclusiveness}.
\end{kitscenes}

\subsection{Choosing the Minimal Data Operator}
\label{subsec:operators}

Once the main research gap is defined, the central engineering decision is \emph{how} to produce the data. We find it useful to think in terms of a small set of \emph{data operators} (\Cref{tab:dataset_operators}) that can be combined almost arbitrarily: \textbf{Exploitation} (adapting, re-labeling, extending, or resampling existing data),
\textbf{Collection} (scraping or mining existing sources), \textbf{Unification} (unifying datasets and APIs), \textbf{Synthesis} (novel data from engine-based simulators or generative models), and \textbf{Recording} (new real-world capture). They differ in cost and output: only recording and synthesis yield genuinely new data, and only recording produces new \emph{real-world} data, at the price of hardware, software, and labeling effort that no other operator incurs.

Our central recommendation follows directly: identify the \emph{minimal} operator, or composition of operators, that solves the diagnosed problem, and treat recording as the operator of last resort. Many impactful datasets did exactly this without any new recording: the synthetic and derived/unified clusters of \Cref{tab:evolution}, including 123D~\cite{Dauner2026123D} unified existing datasets behind a single API, while ArgoTweak~\cite{wild2025argotweak} exploited AV2 map labels for a controlled map-verification benchmark; yet both progressed the field substantially. The cost/benefit of recording is highly context-dependent: it hinges on the infrastructure already available, on the potential for future reuse of the novel real-world data and how broad the enabled research could potentially be, even when focusing mainly on a specific research gap at first.

\newcommand{\y}{\textcolor{green!50!black}{\faCheck}}
\newcommand{\n}{\textcolor{red!70!black}{\faTimes}}
\newcommand{\p}{\textcolor{orange!80!black}{\faAdjust}}

\newcolumntype{L}{>{\raggedright\arraybackslash}p{4.2cm}}

\definecolor{divbg}{gray}{0.92}
\definecolor{headbg}{gray}{0.15}
\definecolor{recbg}{gray}{0.97}

\pagestyle{empty}

\begin{table}[h]
\centering
\renewcommand{\arraystretch}{1.35}
\setlength{\tabcolsep}{10pt}
\small
\resizebox{\textwidth}{!}{%
\begin{tabular}{L  c c c c !{\color{black!40}\vrule width 1.5pt} c}

\toprule

\textbf{} &
\textbf{Exploitation} &
\textbf{Collection} &
\textbf{Unification} &
\textbf{Synthesis} &
\textbf{Recording} \\

\midrule
New (sensor) data   & \n & \n & \n & \y & \y \\
Real-world data     & \y & \y & \y & \n & \y \\

labeling/scenario design required & \n & \y & \n & \y & \y \\
Modify / extend labels & \y & \p & \p & \p & \p \\
Curation / mining required  & \n & \y & \p & \n & \y \\
High compute requirements      & \n & \n & \n & \y & \p \\
Real HW \& recording SW     & \n & \n & \n & \n & \y \\

\makecell[l]{\textit{Notes}} &
\makecell[l]{\footnotesize Relabel or\\[-2pt]\footnotesize modify labels} &
\makecell[l]{\footnotesize Scrape/mine \\[-2pt]\footnotesize or crowdsource } &
\makecell[l]{\footnotesize Unify datasets\\[-2pt]\footnotesize \& APIs} &
\makecell[l]{\footnotesize Sim.\ envs or gen.\ AI;\\[-2pt]\footnotesize scenario/label design} &
\makecell[l]{\footnotesize Sensors \& SW;\\[-2pt]\footnotesize full labeling} \\

\bottomrule
\end{tabular}
}
\caption{ Data Operators vs.\ their properties.
\y~= yes,\ \n~= no,\ \p~= partial.}
\label{tab:dataset_operators}
\end{table}

\begin{kitscenes}
Closing our research gap with recording KITScenes Multimodal was not the only option we discussed. A minimal \emph{Exploitation} plan existed as well: extend Argoverse~2~\cite{Argoverse2} with complete Lanelet2-style 3D relational maps required for deployment-grade HD maps. We abandoned this for three reasons specific to our gap: (i)~the complex European urban road layout we targeted is absent from AV2; (ii)~conclusive long-range evaluation needs sensor and calibration fidelity that AV2's released data cannot provide; and (iii)~reprojection-accurate map labels presuppose the very high-fidelity raw sensor data we lacked. Recording became the minimal viable operator, but was \emph{only} feasible because substantial hardware, recording software, and mapping expertise already existed in-house.

\end{kitscenes}

\subsection{From Operator to a New Dataset: Sensors, Annotation, Scaling and Benchmarks}
\label{subsec:from-operator-to-dataset}

The remaining decisions are largely contingent on the chosen operator. We group them as remaining topics to keep the framework compact; the considerations below apply most directly when recording, but the annotation and scaling guidance generalises to other data operators as well.

\paragraph{Sensor suite design.}
When recording, the sensor suite determines whether new tasks can be enabled or whether a target application can be tackled.
While overlap to existing setups, e.g.\ a central top lidar, allows researchers to compare across domains, novel tasks or saturated benchmarks require changes w.r.t.\ established approaches.
Including both classical and novel systems enables a direct evaluation of the two and supports methods that leverage their unification.
One often overlooked part of the setup is its calibration and synchronization quality, which often determine whether the data can support demanding downstream tasks at all.

\begin{figure}
  \centering
  \includegraphics[width=\linewidth]{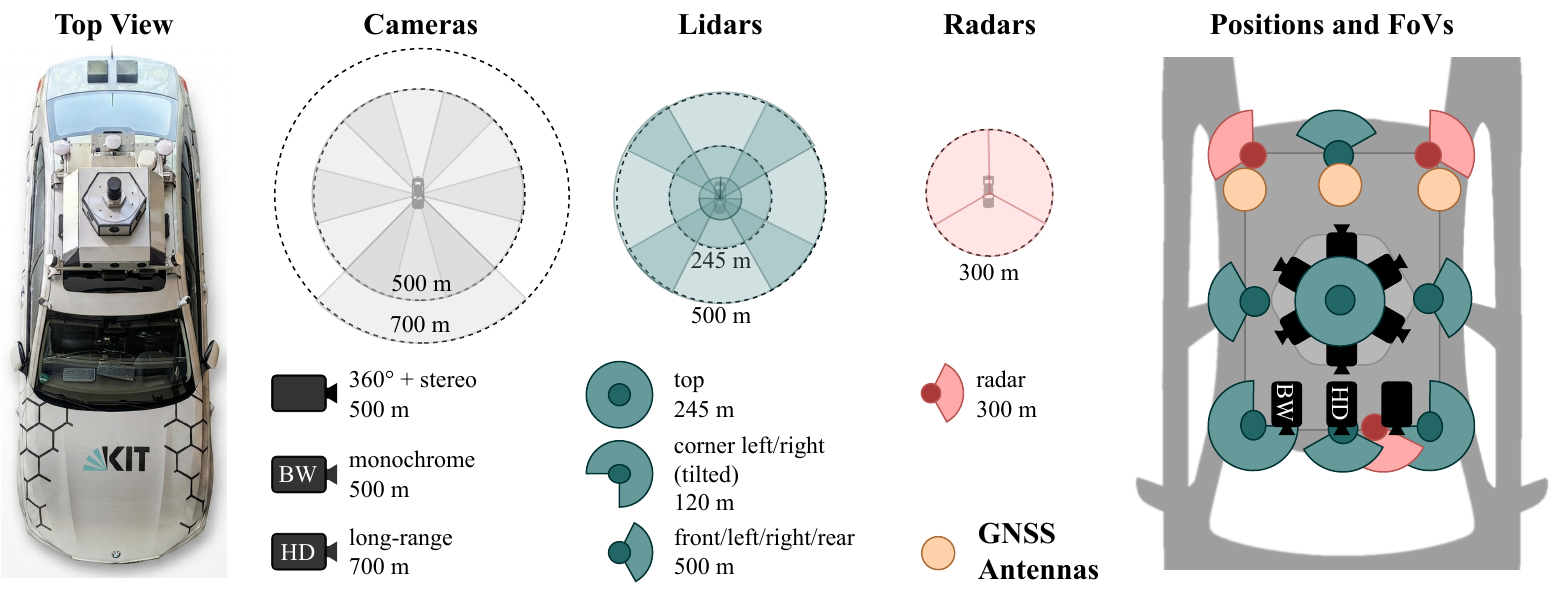}
  \caption{KITScenes Multimodal~\cite{kitscenes_mm} sensor setup. Our sensor rack (left) is depicted along with sensing range (center), as well as sensor positions and their field of view (right).
  }
  \label{fig:sensor_setup}
\end{figure}

\begin{kitscenes}
Our sensor suite combines high-resolution global-shutter cameras, multiple long-range lidars (effective range beyond \SI{400}{\meter}), 4D imaging radar, and redundant GNSS/INS, all hardware-synchronized (\Cref{fig:sensor_setup}).
Choosing a roof-box arrangement over integrated sensors mirrored the L4 robotaxi domain and prioritized fidelity and calibration accuracy over L2 application alignment, reflecting our evaluation-first goal.

\end{kitscenes}

\paragraph{Annotation strategy and quality.}
Annotations can be produced by \emph{human-in-the-loop} labeling (annotators working from guidelines and quality control, aided by automated tools) or by \emph{fully automatic} offline auto-labeling. Human annotation is more costly but higher quality, while at fleet scale the cost of human labeling becomes prohibitive and auto-labeling is the only viable option. While large-scale auto-labeled corpora enable previously unprecedented scaling, curated datasets with the highest calibration, synchronization, sensor, and annotation quality become \emph{more} valuable, precisely because they provide the trustworthy yardstick against which fleet-trained models are measured.

\begin{kitscenes}
To achieve the necessary annotation quality required for evaluation, we follow the human-in-the-loop paradigm, but heavily utilize tooling to simplify the process. For instance, traffic signs or 3D landmarks obtain an initial auto-label~\cite{carnot2026gtsign,pauls2021automatic}, which is then verified by humans. In contrast, HD map lane elements and their topology, including traffic-element-to-lane associations, are labeled manually with tooling support; reliable auto-labeling for maps remains an open problem. We found that quality-testing already-created HD maps for defects and mislabels is effective to automate with heuristics and classification models, leading to a more efficient manual annotation process and a higher quality of the annotations, when combined with a good visualization and UI.
\end{kitscenes}

\paragraph{Getting started and scaling out.}
We strongly recommend defining and producing a small but representatively diverse end-to-end slice of the data and labels as early as possible: a handful of fully processed samples forces the labeling, processing, and evaluation pipelines to exist and exposes the real goal as well as processing issues or hidden assumptions concretely.

\begin{kitscenes}
Even using a small number of samples across cities, sensor setups, calibration versions, or annotation epochs allowed us to reveal a great number of hidden assumptions.
\end{kitscenes}

\paragraph{Benchmarks, baselines, metrics, and leaderboards.}
Three decisions govern whether evaluation is conclusive. First, splits must be free of leakage: in driving data, naive random or temporal splits let geographically adjacent observations bleed across train and test, so separation is best enforced by geographic distance if possible. Second, metrics should be well designed, conclusive, and highly correlated with the application goal. Third, a fully held-out test set served via a leaderboard needs special care in the release, since privileged information (maps, future frames, geo-referenced poses) is easily leaked. Baselines should be easy to reproduce and well established if possible to allow comparison to previous work.

\begin{kitscenes}
KITScenes ships with five initial benchmarks: online HD map construction, long-range monocular depth, novel view synthesis, and end-to-end driving on KITScenes Multimodal, plus long-tail end-to-end driving on KITScenes LongTail. To keep evaluation conclusive, the Multimodal test set withholds all map data (the first held-out test set in online HD map construction) and separates splits by geographic distance rather than time. For application alignment we favor conclusive, map-grounded metrics, e.g.\ the Multi-Maneuver Score~\cite{wagner2026longtail} for driving and relational map-construction metrics that keep detection and topology separable, unlike the aggregated OpenLane-V2 score~\cite{openlaneV2}. We host a public leaderboard for the LongTail benchmark and run a Workshop Challenge to drive fast adoption.
\end{kitscenes}

\section{Outlook and Conclusion}
\label{sec:outlook}

\paragraph{The future of AD research data.}
The training/evaluation divergence of \Cref{sec:evolution} will likely sharpen. On the training side, fleet-scale auto-labeled and generative corpora will keep growing, making the \emph{cheaper} data operators (exploitation, unification, and synthesis) increasingly powerful for teams that cannot record at scale. On the evaluation side, the premium will shift to small, high-fidelity, long-tail sets whose value lies in curation and metric design rather than volume. We expect the most impactful academic contributions to come not from out-recording industry, but from sharply-scoped operators applied to well-diagnosed gaps: new tasks, application-aligned metrics, and conclusive evaluation. Two needs stand out. First, integration with open-source stacks such as Autoware~\cite{autoware} and Apollo, which requires HD maps in standardized formats (Lanelet2~\cite{poggenhans2018lanelet2}, OpenDRIVE~\cite{opendrive}) and the relational annotations, notably 3D traffic lights with explicit lane associations, that rule-based and hybrid planners depend on. Second, the rise of vision-language models opens evaluation of \emph{reasoning} about driving, where detailed HD maps can serve a dual role as geometric priors and as structured grounding for spatial language.

\paragraph{The future of KITScenes.}
Because KITScenes was recorded once at high fidelity, it is naturally a substrate for cheaper downstream operators: its recordings can be \emph{combined} into unified resources such as 123D~\cite{Dauner2026123D}, \emph{exploited} to extend benchmarks in the spirit of OpenLane-V2~\cite{openlaneV2}, and used to \emph{generate} novel views and scenarios for pseudo-closed-loop evaluation. This is the long-term payoff that helped justify recording in the first place: a single high-fidelity capture amortized across many future, low-cost derivatives.

\paragraph{Conclusion.}
Datasets remain foundational to autonomous driving research, but their impact is decided long before any data is collected. We argued that creators should first \emph{diagnose} whether they face a data or an evaluation problem, then \emph{identify} a sharply-defined gap, and only then choose the \emph{minimal data operator} that closes it, treating recording as the operator of last resort, justified by existing infrastructure and downstream reuse.

\clearpage
{
    \footnotesize
    \bibliographystyle{IEEEtran}
    \bibliography{bibliography}

@String(PAMI = {IEEE Trans. Pattern Anal. Mach. Intell.})

@String(CVPR= {IEEE Conf. Comput. Vis. Pattern Recog.})

@String(ICCV= {Int. Conf. Comput. Vis.})

@String(ECCV= {Eur. Conf. Comput. Vis.})

@String(AAAI = {AAAI})

@String(PAMI  = {IEEE TPAMI})

@String(CVPR  = {CVPR})

@String(ICCV  = {ICCV})

@String(ECCV  = {ECCV})

@InProceedings{chen2024maptracker,
author="Chen, Jiacheng
and Wu, Yuefan
and Tan, Jiaqi
and Ma, Hang
and Furukawa, Yasutaka",
editor="Leonardis, Ale{\v{s}}
and Ricci, Elisa
and Roth, Stefan
and Russakovsky, Olga
and Sattler, Torsten
and Varol, G{\"u}l",
title="MapTracker: Tracking with Strided Memory Fusion for Consistent Vector HD Mapping",
booktitle="Computer Vision -- ECCV 2024",
year="2025",
publisher="Springer Nature Switzerland",
address="Cham",
pages="90--107",
isbn="978-3-031-72658-3"
}

@InProceedings{nuscenes,
author = {Caesar, Holger and Bankiti, Varun and Lang, Alex H. and Vora, Sourabh and Liong, Venice Erin and Xu, Qiang and Krishnan, Anush and Pan, Yu and Baldan, Giancarlo and Beijbom, Oscar},
title = {nuScenes: A Multimodal Dataset for Autonomous Driving},
booktitle = {Proceedings of the IEEE/CVF Conference on Computer Vision and Pattern Recognition (CVPR)},
month = {June},
year = {2020}
}

@inproceedings{maptr,
  title     = {MapTR: Structured Modeling and Learning for Online Vectorized HD Map Construction},
  author    = {Liao, Bencheng and Chen, Shaoyu and Wang, Xinggang and Cheng, Tianheng and Zhang, Qian and Liu, Wenyu and Huang, Chang},
  booktitle = {The Eleventh International Conference on Learning Representations},
  year      = {2022}
}

@inproceedings{openlaneV2,
  title     = {Openlane-v2: A topology reasoning benchmark for unified 3d hd mapping},
  author    = {Wang, Huijie and Li, Tianyu and Li, Yang and Chen, Li and Sima, Chonghao and Liu, Zhenbo and Wang, Bangjun and Jia, Peijin and Wang, Yuting and Jiang, Shengyin and others},
  booktitle = {Thirty-seventh Conference on Neural Information Processing Systems Datasets and Benchmarks Track},
  year      = {2023}
}

@inproceedings{Argoverse2,
  author    = {Benjamin Wilson and William Qi and Tanmay Agarwal and John Lambert and Jagjeet Singh and Siddhesh Khandelwal and Bowen Pan and Ratnesh Kumar and Andrew Hartnett and Jhony Kaesemodel Pontes and Deva Ramanan and Peter Carr and James Hays},
  title     = {Argoverse 2: Next Generation Datasets for Self-driving Perception and Forecasting},
  booktitle = {Proceedings of the Neural Information Processing Systems Track on Datasets and Benchmarks (NeurIPS Datasets and Benchmarks 2021)},
  year      = {2021}
}

@inproceedings{poggenhans2018lanelet2,
  author    = {Poggenhans, Fabian and Pauls, Jan-Hendrik and Janosovits, Johannes and Orf, Stefan and Naumann, Maximilian and Kuhnt, Florian and Mayr, Matthias},
  booktitle = {2018 21st International Conference on Intelligent Transportation Systems (ITSC)},
  title     = {Lanelet2: A High-Definition Map Framework for the Future of Automated Driving},
  year      = {2018},
  pages     = {1672--1679},
  doi       = {10.1109/ITSC.2018.8569929},
  address   = {Hawaii, USA},
  month     = {November},
  url       = {http://www.mrt.kit.edu/z/publ/download/2018/Poggenhans2018Lanelet2.pdf}
}

@misc{WaymoE2E,
  author       = "{Waymo Open Dataset}",
  title        = "{Vision-based End-to-End Driving Challenge 2025}",
  year         = 2025,
  url          = {https://waymo.com/open/challenges/2025/e2e-driving},
  note         = {Accessed: 2025-11-01}
}

@misc{NvidiaAD,
  author       = "{PhysicalAI Autonomous Vehicles}",
  title        = "{NVIDIA Autonomous Vehicle Dataset}",
  year         = 2025,
  url          = {https://huggingface.co/datasets/nvidia/PhysicalAI-Autonomous-Vehicles},
  note         = {Accessed: 2026-01-30}
}

@INPROCEEDINGS {Argoverse,
  author = {Ming-Fang Chang and John W Lambert and Patsorn Sangkloy and Jagjeet Singh
       and Slawomir Bak and Andrew Hartnett and De Wang and Peter Carr
       and Simon Lucey and Deva Ramanan and James Hays},
  title = {Argoverse: 3D Tracking and Forecasting with Rich Maps},
  booktitle = {Conference on Computer Vision and Pattern Recognition (CVPR)},
  year = {2019}
}

@InProceedings{WaymoOpenPerception, author = {Sun, Pei and Kretzschmar, Henrik and Dotiwalla, Xerxes and Chouard, Aurelien and Patnaik, Vijaysai and Tsui, Paul and Guo, James and Zhou, Yin and Chai, Yuning and Caine, Benjamin and Vasudevan, Vijay and Han, Wei and Ngiam, Jiquan and Zhao, Hang and Timofeev, Aleksei and Ettinger, Scott and Krivokon, Maxim and Gao, Amy and Joshi, Aditya and Zhang, Yu and Shlens, Jonathon and Chen, Zhifeng and Anguelov, Dragomir}, title = {Scalability in Perception for Autonomous Driving: Waymo Open Dataset}, booktitle = {Proceedings of the IEEE/CVF Conference on Computer Vision and Pattern Recognition (CVPR)}, month = {June}, year = {2020} }

@InProceedings{WaymoOpenMotion, author={Ettinger, Scott and Cheng, Shuyang and Caine, Benjamin and Liu, Chenxi and Zhao, Hang and Pradhan, Sabeek and Chai, Yuning and Sapp, Ben and Qi, Charles R. and Zhou, Yin and Yang, Zoey and Chouard, Aur'elien and Sun, Pei and Ngiam, Jiquan and Vasudevan, Vijay and McCauley, Alexander and Shlens, Jonathon and Anguelov, Dragomir}, title={Large Scale Interactive Motion Forecasting for Autonomous Driving: The Waymo Open Motion Dataset}, booktitle= {Proceedings of the IEEE/CVF International Conference on Computer Vision (ICCV)}, month={October}, year={2021}, pages={9710-9719} }

@INPROCEEDINGS{Kitti,
  author={Geiger, Andreas and Lenz, Philip and Urtasun, Raquel},
  booktitle={2012 IEEE Conference on Computer Vision and Pattern Recognition},
  title={Are we ready for autonomous driving? The KITTI vision benchmark suite},
  year={2012},
  volume={},
  number={},
  pages={3354-3361},
  doi={10.1109/CVPR.2012.6248074}}

@INPROCEEDINGS{nuPlan,
  author={Karnchanachari, Napat and Geromichalos, Dimitris and Tan, Kok Seang and Li, Nanxiang and Eriksen, Christopher and Yaghoubi, Shakiba and Mehdipour, Noushin and Bernasconi, Gianmarco and Fong, Whye Kit and Guo, Yiluan and Caesar, Holger},
  booktitle={2024 IEEE International Conference on Robotics and Automation (ICRA)},
  title={Towards learning-based planning: The nuPlan benchmark for real-world autonomous driving},
  year={2024},
  volume={},
  number={},
  pages={629-636},
  doi={10.1109/ICRA57147.2024.10610077}}

@inproceedings{virtualkitti,
    author = {Gaidon, A and Wang, Q and Cabon, Y and Vig, E},
    title = {Virtual Worlds as Proxy for Multi-Object Tracking Analysis},
    booktitle = {CVPR},
    year = {2016}
}

@inproceedings{wild2025argotweak,
  title={ArgoTweak: Towards Self-Updating HD Maps through Structured Priors},
  author={Lena Wild and Rafael Valencia and Patric Jensfelt},
  booktitle={Proceedings of the IEEE/CVF International Conference on Computer Vision (ICCV)},
  year={2025}
}

@inproceedings{jia2024bench2drive,
  title={Bench2Drive: Towards Multi-Ability Benchmarking of Closed-Loop End-To-End Autonomous Driving},
  author={Xiaosong Jia and Zhenjie Yang and Qifeng Li and Zhiyuan Zhang and Junchi Yan},
  booktitle={NeurIPS 2024 Datasets and Benchmarks Track},
  year={2024}
}

@article{Dauner2026123D,
  title={123D: Unifying Multi-Modal Autonomous Driving Data at Scale},
  author={Dauner, Daniel and Charraut, Valentin and Berle, Bastian and Li, Tianyu and Nguyen, Long and Wang, Jiabao and Jing, Changhui and Igl, Maximilian and Caesar, Holger and Ivanovic, Boris and Geiger, Andreas and Chitta, Kashyap},
  journal={arXiv preprint arXiv:2605.08084},
  year={2026}
}

@inproceedings{immel2026sdtagnet,
  title     = {{SDT}agNet: Leveraging Text-Annotated Navigation Maps for Online {HD} Map Construction},
  author    = {Fabian Immel and Jan-Hendrik Pauls and Richard Fehler and Frank Bieder and Jonas Merkert and Christoph Stiller},
  booktitle = {The Thirty-ninth Annual Conference on Neural Information Processing Systems},
  year      = {2025},
  url       = {https://openreview.net/forum?id=N3E1cU8Cv3}
}

@inproceedings{Cordts2016Cityscapes,
  title     = {The Cityscapes Dataset for Semantic Urban Scene Understanding},
  author    = {Cordts, Marius and Omran, Mohamed and Ramos, Sebastian and Rehfeld, Timo and Enzweiler, Markus and Benenson, Rodrigo and Franke, Uwe and Roth, Stefan and Schiele, Bernt},
  booktitle = {Proc. of the IEEE Conference on Computer Vision and Pattern Recognition (CVPR)},
  year      = {2016}
}

@manual{opendrive,
  title  = {{ASAM OpenDRIVE 1.8.0 Specification}},
  author = {{ASAM e.V.}},
  year   = {2023},
  month  = {November},
  url    = {https://www.asam.net/standards/detail/opendrive/},
  note   = {Published November 22, 2023}
}

@inproceedings{mao2021once,
  author    = {Mao, Jiageng and Minzhe, Niu and Jiang, ChenHan and liang, hanxue and Chen, Jingheng and Liang, Xiaodan and Li, Yamin and Ye, Chaoqiang and Zhang, Wei and Li, Zhenguo and Yu, Jie and XU, Chunjing and Xu, Hang},
  booktitle = {Proceedings of the Neural Information Processing Systems Track on Datasets and Benchmarks},
  editor    = {J. Vanschoren and S. Yeung},
  pages     = {},
  title     = {One Million Scenes for Autonomous Driving: ONCE Dataset},
  url       = {https://datasets-benchmarks-proceedings.neurips.cc/paper_files/paper/2021/file/67c6a1e7ce56d3d6fa748ab6d9af3fd7-Paper-round1.pdf},
  volume    = {1},
  year      = {2021}
}

@article{huang2018apolloscape,
  author    = {Huang, Xinyu and Wang, Peng and Cheng, Xinjing and Zhou, Dingfu and Geng, Qichuan and Yang, Ruigang},
  journal   = { IEEE Transactions on Pattern Analysis \& Machine Intelligence },
  title     = {{ The ApolloScape Open Dataset for Autonomous Driving and Its Application }},
  year      = {2020},
  volume    = {42},
  number    = {10},
  issn      = {1939-3539},
  pages     = {2702-2719},
  doi       = {10.1109/TPAMI.2019.2926463},
  url       = {https://doi.ieeecomputersociety.org/10.1109/TPAMI.2019.2926463},
  publisher = {IEEE Computer Society},
  address   = {Los Alamitos, CA, USA},
  month     = oct
}

@article{Liao2022PAMI,
  title   = {{KITTI}-360: A Novel Dataset and Benchmarks for Urban Scene Understanding in 2D and 3D},
  author  = {Yiyi Liao and Jun Xie and Andreas Geiger},
  journal = {Pattern Analysis and Machine Intelligence (PAMI)},
  year    = {2022}
}

@article{maptrv2,
  author   = {Liao, Bencheng
              and Chen, Shaoyu
              and Zhang, Yunchi
              and Jiang, Bo
              and Zhang, Qian
              and Liu, Wenyu
              and Huang, Chang
              and Wang, Xinggang},
  title    = {MapTRv2: An End-to-End Framework for Online Vectorized HD Map Construction},
  journal  = {International Journal of Computer Vision},
  year     = {2024},
  month    = {Oct},
  day      = {06},
  issn     = {1573-1405},
  doi      = {10.1007/s11263-024-02235-z},
  url      = {https://doi.org/10.1007/s11263-024-02235-z}
}

@inproceedings{alibeigi2023zenseact,
  title     = {Zenseact Open Dataset: A large-scale and diverse multimodal dataset for autonomous driving},
  author    = {Alibeigi, Mina and Ljungbergh, William and Tonderski, Adam and Hess, Georg and Lilja, Adam and Lindstrom, Carl and Motorniuk, Daria and Fu, Junsheng and Widahl, Jenny and Petersson, Christoffer},
  booktitle = {Proceedings of the IEEE/CVF International Conference on Computer Vision},
  year      = {2023}
}

@misc{autoware,
  title        = {{Autoware}},
  author       = {{Autoware Foundation}},
  howpublished = {\url{https://github.com/autowarefoundation/autoware}},
  note         = {Accessed: 2026-05-02}
}

@inproceedings{pauls2021automatic,
  author    = {Pauls, Jan-Hendrik and Schmidt, Benjamin and Stiller, Christoph},
  booktitle = {2021 IEEE International Conference on Robotics and Automation (ICRA)},
  title     = {Automatic Mapping of Tailored Landmark Representations for Automated Driving and Map Learning},
  year      = {2021},
  pages     = {6725-6731},
  doi       = {10.1109/ICRA48506.2021.9561432}
}

@inproceedings{carnot2026gtsign,
  author    = {Carnot, Miriam Louise and Fastermann, Erik and Kunze, Jonas and Peukert, Eric and Ludwig, André and Franczyk, Bogdan},
  title     = {GTSIGN-220: A Crowd-Sourced, StVO-Aligned Benchmark for Fine-Grained German Traffic Sign Recognition},
  booktitle = {Intelligent Vehicles Symposium (IV)},
  year      = {2026}
}

@inproceedings{yuan2024streammapnet,
  title     = {Streammapnet: Streaming mapping network for vectorized online hd map construction},
  author    = {Yuan, Tianyuan and Liu, Yicheng and Wang, Yue and Wang, Yilun and Zhao, Hang},
  booktitle = {Proceedings of the IEEE/CVF Winter Conference on Applications of Computer Vision},
  pages     = {7356--7365},
  year      = {2024}
}

@misc{wagner2026longtail,
  title         = {LongTail Driving Scenarios with Reasoning Traces: The KITScenes LongTail Dataset},
  author        = {Royden Wagner and Omer Sahin Tas and Jaime Villa and Felix Hauser and Yinzhe Shen and Marlon Steiner and Dominik Strutz and Carlos Fernandez and Christian Kinzig and Guillermo S. Guitierrez-Cabello and Hendrik Königshof and Fabian Immel and Richard Schwarzkopf and Nils Alexander Rack and Kevin Rösch and Kaiwen Wang and Jan-Hendrik Pauls and Martin Lauer and Igor Gilitschenski and Holger Caesar and Christoph Stiller},
  year          = {2026},
  eprint        = {2603.23607},
  archiveprefix = {arXiv},
  primaryclass  = {cs.CV},
  url           = {https://arxiv.org/abs/2603.23607}
}

@inproceedings{wang2024stream_sqd_mapnet,
  title={Stream query denoising for vectorized hd-map construction},
  author={Wang, Shuo and Jia, Fan and Mao, Weixin and Liu, Yingfei and Zhao, Yucheng and Chen, Zehui and Wang, Tiancai and Zhang, Chi and Zhang, Xiangyu and Zhao, Feng},
  booktitle={European Conference on Computer Vision},
  pages={203--220},
  year={2024},
  organization={Springer}
}

@article{shi2024globalmapnet,
  title={Globalmapnet: An online framework for vectorized global hd map construction},
  author={Shi, Anqi and Cai, Yuze and Chen, Xiangyu and Pu, Jian and Fu, Zeyu and Lu, Hong},
  journal={arXiv preprint arXiv:2409.10063},
  year={2024}
}

@inproceedings{zhang2024enhancing_HR_mapnet,
  title={Enhancing vectorized map perception with historical rasterized maps},
  author={Zhang, Xiaoyu and Liu, Guangwei and Liu, Zihao and Xu, Ningyi and Liu, Yunhui and Zhao, Ji},
  booktitle={European Conference on Computer Vision},
  pages={422--439},
  year={2024},
  organization={Springer}
}

@inproceedings{zhang2025mapexpert,
  title={Mapexpert: Online hd map construction with simple and efficient sparse map element expert},
  author={Zhang, Dapeng and Chen, Dayu and Zhi, Peng and Chen, Yinda and Yuan, Zhenlong and Li, Chenyang and Zhou, Rui and Zhou, Qingguo and others},
  booktitle={Proceedings of the AAAI Conference on Artificial Intelligence},
  volume={39},
  number={14},
  pages={14745--14753},
  year={2025}
}

@article{yang2025histrackmap,
  title={Histrackmap: Global vectorized high-definition map construction via history map tracking},
  author={Yang, Jing and Yang, Sen and Tan, Xiao and Wang, Hanli},
  journal={arXiv preprint arXiv:2503.07168},
  year={2025}
}

@article{erdougan2025mapping_skeptic,
  title={Mapping like a Skeptic: Probabilistic BEV Projection for Online HD Mapping},
  author={Erdo{\u{g}}an, Fatih and Bar{\i}n, Merve Rabia and G{\"u}ney, Fatma},
  journal={arXiv preprint arXiv:2508.21689},
  year={2025}
}

@misc{nvidia2025cosmosdrivedreams,
  title  = {Cosmos-Drive-Dreams: Scalable Synthetic Driving Data Generation with World Foundation Models},
  author = {Ren, Xuanchi and Lu, Yifan and Cao, Tianshi and Gao, Ruiyuan and
            Huang, Shengyu and Sabour, Amirmojtaba and Shen, Tianchang and
            Pfaff, Tobias and Wu, Jay Zhangjie and Chen, Runjian and
            Kim, Seung Wook and Gao, Jun and Leal-Taixe, Laura and
            Chen, Mike and Fidler, Sanja and Ling, Huan},
  year   = {2025},
  url    = {https://arxiv.org/abs/2506.09042}
}

@misc{kitscenes_mm,
  title     = {The Road Ahead in Autonomous Driving: The KITScenes Multimodal Dataset},
  author    = {Richard Schwarzkopf and Fabian Immel and Alexander Blumberg and Jonas Merkert and Nils Rack and Kaiwen Wang and
               Fabian Konstantinidis and Julian Truetsch and Carlos Fernandez and Annika Bätz and Kevin Rösch and Marlon Steiner and
               Willi Poh and Yinzhe Shen and Royden Wagner and Felix Hauser and Dominik Strutz and Jaime Villa and Gleb Stepanov and
               Holger Caesar and Ömer Şahin Taş and Frank Bieder and Jan-Hendrik Pauls and Christoph Stiller},
  year      = {2026},
  eprint    = {2606.02956},
  archivePrefix = {arXiv},
  primaryClass  = {cs.CV},
  url       = {https://arxiv.org/abs/2606.02956}
}

@inproceedings{behley2019iccv,
  author    = {Behley, Jens and Garbade, Martin and Milioto, Andres and Quenzel, Jan and Behnke, Sven and Stachniss, Cyrill and Gall, Juergen},
  title     = {{SemanticKITTI}: A Dataset for Semantic Scene Understanding of {LiDAR} Sequences},
  booktitle = {Proceedings of the IEEE/CVF International Conference on Computer Vision (ICCV)},
  year      = {2019}
}

@inproceedings{dosovitskiy2017carla,
  author    = {Dosovitskiy, Alexey and Ros, German and Codevilla, Felipe and Lopez, Antonio and Koltun, Vladlen},
  title     = {{CARLA}: An Open Urban Driving Simulator},
  booktitle = {Proceedings of the 1st Annual Conference on Robot Learning (CoRL)},
  pages     = {1--16},
  volume    = {78},
  series    = {Proceedings of Machine Learning Research},
  publisher = {PMLR},
  year      = {2017}
}

@inproceedings{bdd100k,
  author    = {Yu, Fisher and Chen, Haofeng and Wang, Xin and Xian, Wenqi and Chen, Yingying and Liu, Fangchen and Madhavan, Vashisht and Darrell, Trevor},
  title     = {{BDD100K}: A Diverse Driving Dataset for Heterogeneous Multitask Learning},
  booktitle = {Proceedings of the IEEE/CVF Conference on Computer Vision and Pattern Recognition (CVPR)},
  year      = {2020}
}

@inproceedings{neuhold2017mapillary,
  author    = {Neuhold, Gerhard and Ollmann, Tobias and Rota Bul\`{o}, Samuel and Kontschieder, Peter},
  title     = {The Mapillary Vistas Dataset for Semantic Understanding of Street Scenes},
  booktitle = {Proceedings of the IEEE International Conference on Computer Vision (ICCV)},
  pages     = {5000--5009},
  year      = {2017},
  doi       = {10.1109/ICCV.2017.534}
}

@misc{schafer2018comma2k19,
  author    = {Schafer, Harald and Santana, Eder and Haden, Andrew and Biasini, Riccardo},
  title     = {A Commute in Data: The comma2k19 Dataset},
  year      = {2018},
  eprint    = {1812.05752},
  archivePrefix = {arXiv},
  primaryClass  = {cs.RO}
}

@inproceedings{houston2021lyft,
  author    = {Houston, John and Zuidhof, Guido and Bergamini, Luca and Ye, Yawei and Chen, Long and Jain, Ashesh and Omari, Sammy and Iglovikov, Vladimir and Ondruska, Peter},
  title     = {One Thousand and One Hours: Self-driving Motion Prediction Dataset},
  booktitle = {Proceedings of the 2020 Conference on Robot Learning (CoRL)},
  pages     = {409--418},
  volume    = {155},
  series    = {Proceedings of Machine Learning Research},
  publisher = {PMLR},
  year      = {2021}
}

@inproceedings{pandaset2021,
  author    = {Xiao, Pengchuan and Shao, Zhenlei and Hao, Steven and Zhang, Zishuo and Chai, Xiaolin and Jiao, Judy and Li, Zesong and Wu, Jian and Sun, Kai and Jiang, Kun and Wang, Yunlong and Yang, Diange},
  title     = {{PandaSet}: Advanced Sensor Suite Dataset for Autonomous Driving},
  booktitle = {2021 IEEE International Intelligent Transportation Systems Conference (ITSC)},
  pages     = {3095--3101},
  year      = {2021},
  doi       = {10.1109/ITSC48978.2021.9565009}
}

@misc{geyer2020a2d2,
  author    = {Geyer, Jakob and Kassahun, Yohannes and Mahmudi, Mentar and Ricou, Xavier and Durgesh, Rupesh and Chung, Andrew S. and Hauswald, Lorenz and Pham, Viet Hoang and M\"{u}hlegg, Maximilian and Dorn, Sebastian and Fernandez, Tiffany and J\"{a}nicke, Martin and Mirashi, Sudesh and Savani, Chiragkumar and Sturm, Martin and Vorobiov, Oleksandr and Oelker, Martin and Garreis, Sebastian and Schuberth, Peter},
  title     = {{A2D2}: Audi Autonomous Driving Dataset},
  year      = {2020},
  eprint    = {2004.06320},
  archivePrefix = {arXiv},
  primaryClass  = {cs.CV},
  url       = {https://www.a2d2.audi}
}

@inproceedings{Dauner2024NEURIPS,
  author    = {Dauner, Daniel and Hallgarten, Marcel and Li, Tianyu and Weng, Xinshuo and Huang, Zhiyu and Yang, Zetong and Li, Hongyang and Gilitschenski, Igor and Ivanovic, Boris and Pavone, Marco and Geiger, Andreas and Chitta, Kashyap},
  title     = {{NAVSIM}: Data-Driven Non-Reactive Autonomous Vehicle Simulation and Benchmarking},
  booktitle = {Advances in Neural Information Processing Systems (NeurIPS)},
  year      = {2024}
}

@inproceedings{ljungbergh2024neuroncap,
  author    = {Ljungbergh, William and Tonderski, Adam and Johnander, Joakim and Caesar, Holger and {\AA}str\"{o}m, Kalle and Felsberg, Michael and Petersson, Christoffer},
  title     = {{NeuroNCAP}: Photorealistic Closed-loop Safety Testing for Autonomous Driving},
  booktitle = {Proceedings of the European Conference on Computer Vision (ECCV)},
  pages     = {161--177},
  year      = {2024},
  publisher = {Springer}
}
}

\end{document}